# Is GPT a Computational Model of Emotion? Detailed Analysis

Ala N. Tak and Jonathan Gratch
Institute for Creative Technologies
University of Southern California
Playa Vista, CA 90094, USA.
nekouvag@usc.edu, gratch@ict.usc.edu

## Contents

## Abstract

This paper investigates the emotional reasoning abilities of the GPT family of large language models via a component perspective. The paper first examines how the model reasons about autobiographical memories. Second, it systematically varies aspects of situations to impact emotion intensity and coping tendencies. Even without the use of prompt engineering, it is shown that GPT's predictions align significantly with human-provided appraisals and emotional labels. However, GPT faces difficulties predicting emotion intensity and coping responses. GPT-4 showed the highest performance in the initial study but fell short in the second, despite providing superior results after minor prompt engineering. This assessment brings up questions on how to effectively employ the strong points and address the weak areas of these models, particularly concerning response variability. These studies underscore the merits of evaluating models from a componential perspective [1].

# 1. Study 1

## 1.1 Original prompts

GPT is sensitive to minor variations in prompt design [2]. To mitigate this, we adopt the strategy of Binz and Schulz to evaluate GPT's cognitive reasoning capabilities [3]. We prompt the model (without any fine-tuning) with the exact question pattern used for human respondents in a psychological experiment, appending only the least required additional text to enable the model to produce uniform answers, like responding to Likert scales. Figure SM.1 is the exact prompt given to GPT in Study 1.

Each prompt is provided in a separate conversation, and the text in red is replaced with each story. This was run once for each story. The model occasionally returned additional explanatory text along with each response which was ignored for analysis. The model always returned four emotion labels. The appraisal items are verbatim from Smith and Lazarus 1990 [4].

> Imagine the following hypothetical situation and answer the following questions. This is the situation:
> "My grandfather passed away a few weeks prior owing to health complications. Processing the event in the absence of family was a tall task. The fact that he was doing well a few months back when I was leaving for the USA and the sudden demise were things which compounded the impact of the event. It took a few weeks for me to return to state of normalcy, process the entire event and accepting it."
> Please rate the situation along the following criteria from 1 to 9. Only provide a number
> 1. **Relevance**: How important was what was happening in the situation to you? (1: not at all … 9: extremely)
> 2. **Congruence**: Think about what you did and did not want in this situation. How consistent was the situation with what you wanted? (1: not at all … 9: extremely)
> 3. **Self-accountability**: To what extent did you consider YOURSELF responsible for the situation? (1: not at all … 9: extremely)
> 4. **Other-accountability**: To what extent did you consider SOMEONE ELSE responsible for the situation? (1: not at all … 9: extremely)
> 5. **Future-expectancy**: Think about how you wanted this situation to turn out. How consistent with these wishes did you expect the situation to become (or stay)? (1: not at all … 9: extremely)
> 6. **Problem-focused coping**: Think about what you did and didn't want in this situation. How certain were you that you would be able to influence things to make (or keep) the situation the way you wanted it? (1: certainly WILL not be able … certainly WILL be able)
> 7. **Accommodative-focused coping**: How certain were you that you would be able to deal emotionally with what was happening in this situation? (1: not able to cope … 9: completely able to cope)
> 8. Finally, please list at most four emotions someone in this situation is likely to feel.
>
> **Figure SM.1: Prompt used in Study 1.**

## 1.2 Emotion derivation

Human participants offered from one to eight emotional labels for their stories (M=2.31, SD=1.39). GPT-3.5 and GPT-4 always returned four labels. We explored two general approaches for comparing these labels. First, as reported in the paper [5], we converted labels into valence, arousal, and dominance scores. The results in the paper use a dictionary-based method as people reported very common emotion terms like joy, anger, or disappointment. We also complement this with an embedding approach summarized here. Second, we examined if one of the words output by GPT was an exact match for one of the words provided by the participant, where different grammatical forms of the identical word were considered a match (e.g., angry matches anger, but fear does not match scared). Interestingly, the first word reported by GPT was the best match, suggesting that the first word provided by the model is its best guess.

The dictionary results are reported in the paper. Here we report the embedding and word-match results.

### 1.2.1 Embedding results

We approach this problem using word embeddings, such as those provided by Word2Vec, combined with distance/similarity metrics, such as cosine similarity. Word embeddings represent words in a multi-dimensional space and are generated in such a way that similar words are close to each other in this space. We first take each pair of emotion labels, calculate their word vectors (using Word2Vec [6]), and then measure the cosine similarity between the vectors. Our analysis reveals an average general similarity of approximately 0.66 and 0.50 across all comparisons using GPT-3.5 and GPT-4 output, respectively, indicating moderate-to-strong similarity. This approach assumes that similar word embeddings would have similar emotional content, which is a simplification. Word embeddings capture many facets of a word's meaning, which includes but is not limited to its emotional content. As a result, while the cosine similarity of word embeddings can serve as a rough proxy for emotional similarity, it will not fully capture the valence and arousal dimensions.

To discover certain "directions" in the word embedding space that seem to correspond to particular semantic differences (i.e., emotional content), we projected word vectors onto the "VAD" dimension in Word2Vec and compared the labels in terms of this projection. However, Word2Vec does not inherently have an interpretable VAD dimension. Thus, we identified pairs of words that differ mainly in terms of V (or A, D) and subtracted their vectors to find the difference vectors. We average these difference vectors to find a vector that roughly points in the "V" (or A, D) direction in the word embedding space. Finally, we computed the correlation between the projections of GPT and human labels to the generated VAD directions, which is presented in Table SM.1.

| Table SM.1 | Correlation with human-reported emotion | | |
|---|---|---|---|
| **Models** | **Valence** | **Arousal** | **Dominance** |
| GPT-3.5 | r = 0.793, p < .001*** | r = 0.690, p < .001*** | r = 0.337, p=.044 |
| GPT-4 | r = 0.779, p < .001*** | r = 0.532, p < .001*** | r = 0.026, p=.881 |

It should be noted that this method assumes that the difference vectors capture the semantic difference between words as intended, which is not always true. Also, we assume that the "V" (or A, D) dimension is orthogonal to the other dimensions in the word embedding space, which may not be the case. Lastly, the choice of word pairs can greatly affect the resulting VAD vectors.

### 1.2.2 Word-match results

Table SM.2 lists how often a GPT-provided label matches one of the human-provided emotion labels. This is broken out by the order of words produced by the model. For example, the first label provided by GPT-3.5 matched one of the human-provided labels for a given story 42.9% of the time. The second label only matched 34.3% of the time, and so forth. Overall, at least one of the labels matched at least one of the human responses 80% of the time. GPT-4 was slightly less accurate than GPT-3.5 on this metric, but this difference failed to reach significance: $\chi^2$ (1, N = 35) = 0.8, p = .771.

| Table SM.2 | Position of GPT-reported label | | | | |
|---|---|---|---|---|---|
| Model | First | Second | Third | Fourth | Any |
| GPT-3.5 | 0.429 | 0.343 | 0.257 | 0.171 | 0.800 |
| GPT-4 | 0.371 | 0.343 | 0.314 | 0.114 | 0.771 |

## 1.3 Affect derivation

Appraisal derivation considers which appraisals predict specific emotions. As people reported multiple emotion labels, we predict the average valence, arousal, and dominance scores associated with each story. Thus, we performed backward linear regression separately to predict average valence, average arousal, and average dominance. This is first performed on human data and then on model data. Figure 5 illustrates the results for GPT4. Figure SM.2 shows the results for GPT3.5.

Appraisal theory claims the valence of responses is dictated by if the situation is goal-congruent. This is indeed the association found in the human data but GPT-3 primarily associates valence with future-expectancy (which refers to if the situation unfolded as expected). Through post hoc analysis, this seems to arise due to collinearities between GPT-3's interpretation of goal-congruence and future expectancy that are less present in human ratings.

Appraisal theory claims arousal should largely be determined by the relevance of the event to the individual (e.g., a threat to a very important goal would be more relevant than a threat to a minor goal). This is indeed the association found in the human data, but GPT associates arousal with other-accountability, though it should be noted that both associations are weak.

Finally, appraisal theory claims dominance should be associated with perceptions of control (positively associated with problem-focused coping and negatively associated with emotion-focused coping). Neither of these associations was found in either model. Self-reported dominance was associated with goal-congruence, which makes some sense as people are presumably more in control in positive situations. GPT-3 associates dominance with future expectancy, likely for the same reasons it uses this feature for valence.

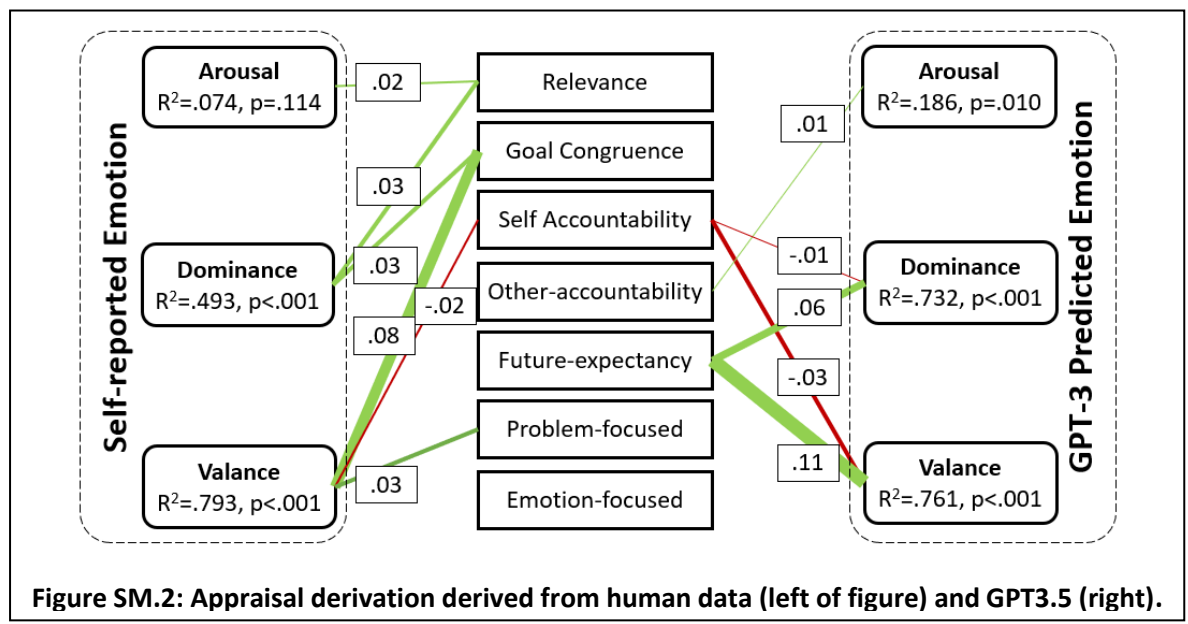

**Figure SM.2: Appraisal derivation derived from human data (left of figure) and GPT3.5 (right).**

# 2. Study 2

## 2.1 Original prompts

*2.1.1 Prompt*

In this section, we provide a complete report of the prompts queried for each state in the game (Lost, Losing, Start, Winning, Won) and three utility levels (Low, Medium, and High) using the coin flip game adapted from [7].

For utility in ('a little bit', 'a medium amount', 'a lot'):

> For game_state in ('Imagine you have lost the game.', 'Imagine you have obtained zero heads and your opponent has two heads. It is your turn.', 'Imagine the game is just about to start. It is your turn.', 'Imagine you have obtained two heads and your opponent has zero heads. It is your turn.', 'Imagine you have won the game.')
>
> > *"Imagine you are playing a simple game with someone. \n"*
> > *"You take turns flipping a coin. The first person to get four heads wins. \n"*
>
> If lost or won:
> > *"{game_state} Imagine that you were playing for {utility} of money.\n"*
>
> Else:
> > *"{game_state} Imagine that you are playing for {utility} of money.\n"*
>
> > *"Answer the following questions by providing only a single number on a scale from 1 to 100, where 1 means not at all and 100 means a lot: 1. How important to you is it that you win? 2. How likely is it that you win? 3. How much control do you have over winning? 4. How much do you feel hope? 5. How much do you feel fear? 6. How much do you feel joy? 7. How much do you feel sadness? 8. How much do you feel anger? \n"*
> > *"Please do not respond anything else other than the answers to the 8 questions above.\n"*
> > *"Please put the answer in the following JSON format and make all data types to be string and use all lowercase. It is very important.\n"*
>
> > *'{"1":" ", "2":" ", "3": "", "4": "", "5": "", "6": "", "7": "", "8": ""}\n'*

*2.1.1 Results*

Figure SM.3 demonstrates emotion intensity from human self-report compared with GPT in response to different states of the coin-flip game. Intensity is on the y-axis, whereas reported probability of winning the game is reported on the x-axis. GPT graphs show 95% confidence intervals of the mean.

Based on the two-way ANOVA conducted on the four dependent variables (hope, fear, joy, and sadness), the main effects of relevance and game state, as well as the interaction effect between relevance and game state, as well as partial eta squared ($\eta^2$) values, 95% confidence interval (CI), are summarized in Table SM.3.

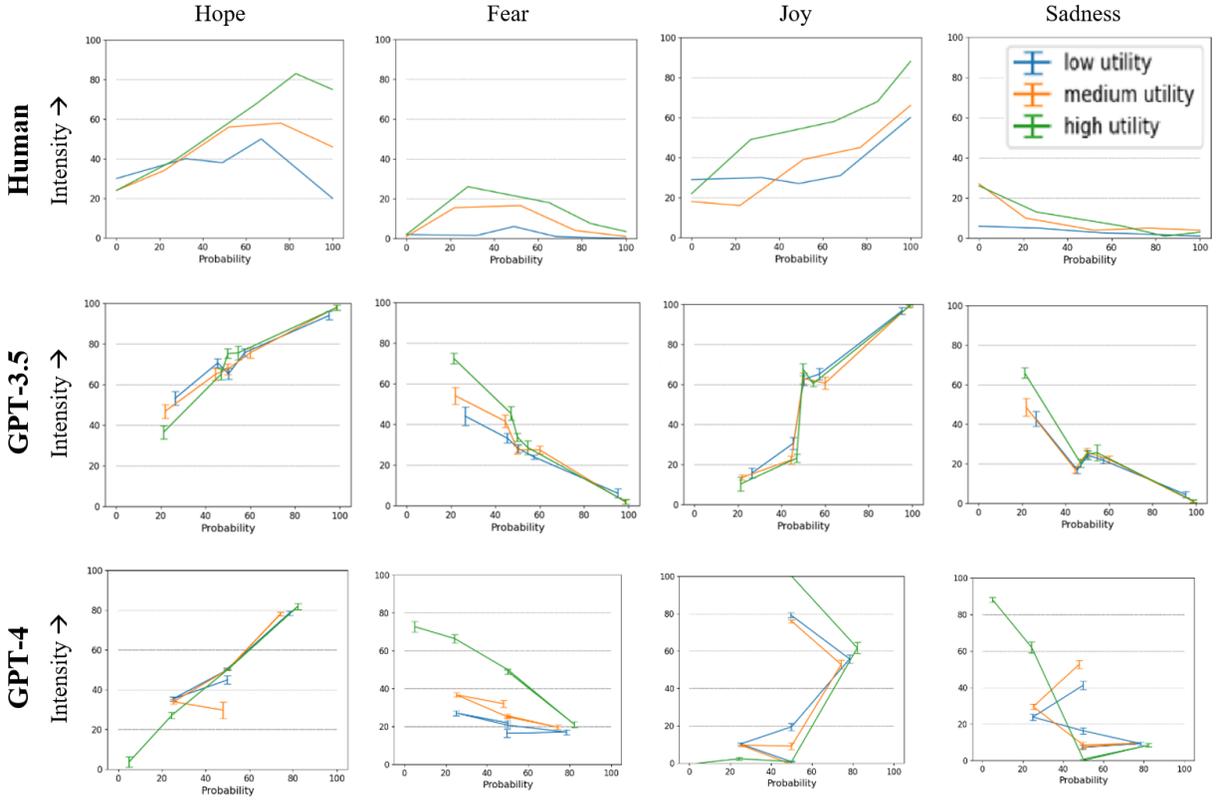

**Figure SM.3:** Intensity derivation results (corresponding to Fig 8. in the paper)

| Table SM.3 | | Impact of game state and goal-relevance for each emotion | | |
|---|---|---|---|---|
| | | *Goal-relevance* | *Game State* | *Interaction Effect* |
| GPT-3.5 | Hope | F(2, 1485) = 2.15, p = 0.117, η² = .003 | F(4, 1485) = 579.34, p < .001***, η² = .61 | F(8, 1485) = 15.49, p < .001***, η² = .08 |
| | Fear | F(2, 1485) = 62.44, p < .001***, η² = .08 | F(4, 1485) = 645.67, p < .001***, η² = .63 | F(8, 1485) = 21.81, p < .001***, η² = .11 |
| | Joy | F(2, 1485) = 5.98, p = .002***, η² = .008 | F(4, 1485) = 2409.07, p < .001***, η² = .87 | F(8, 1485) = 6.34, p < .001***, η² = .03 |
| | Sadness | F(2, 1485) = 30.27, p < .001***, η² = .04 | F(4, 1485) = 691.91, p < .001***, η² = .65 | F(8, 1485) = 19.25, p < .001***, η² = .09 |
| GPT-4 | Hope | F(2, 1485) = 173.0, p < .001***, η² = .19 | F(4, 1485) = 2035.9, p < .001***, η² = .85 | F(8, 1485) = 135.6, p < .001***, η² = .42 |
| | Fear | F(2, 1485) = 2241.8, p < .001***, η² = .75 | F(4, 1485) = 490.0, p < .001***, η² = .57 | F(8, 1485) = 143.2, p < .001***, η² = .44 |
| | Joy | F(2, 1485) = 39.67, p < .001***, η² = .05 | F(4, 1485) = 8182.93, p < .001***, η² = .96 | F(8, 1485) = 136.81, p < .001***, η² = .42 |
| | Sadness | F(2, 1485) = 364, p < .001***, η² = .33 | F(4, 1485) = 3001, p < .001***, η² = .89 | F(8, 1485) = 369, p < .001***, η² = .67 |

Figure SM.4 illustrates emotional distancing/engagement from the goal of winning as a function of the game state. The left shows human self-report, and the middle and right are predictions from GPT models. Both models fail to predict engagement.

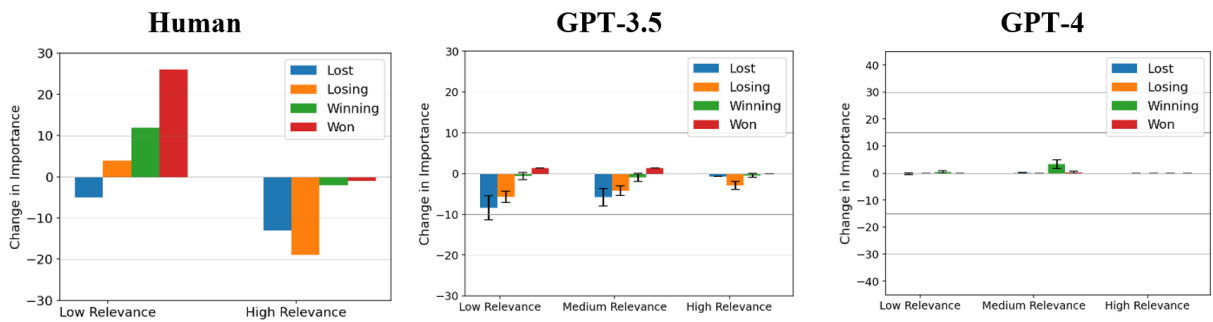

**Figure SM.4: Consequence derivation results (corresponding to Fig 9. in the paper)**

ANOVA results show that there are significant main effects of relevance and game state, as well as a significant interaction effect between them on importance. Table SM.4 provides a summary of the results.

| Table SM.4 (Table 4 in the paper) | | Impact of game state and relevance on importance of winning | | |
|---|---|---|---|---|
| | | *F value* | *p* | *η² (partial)* |
| GPT-3.5 | Goal-relevance | 41.73 | $p < .001$*** | 0.05 |
| GPT-3.5 | Game State | 59.55 | $p < .001$*** | 0.14 |
| GPT-3.5 | Interaction Effect | 9.85 | $p < .001$*** | 0.05 |
| GPT-4 | Goal-relevance | 78091.57 | $p < .001$*** | 0.99 |
| GPT-4 | Game State | 17.05 | $p < .001$*** | 0.04 |
| GPT-4 | Interaction Effect | 12.10 | $p < .001$*** | 0.06 |

## 2.2 Prompt engineering

### 2.2.1 Prompt

We applied incremental adjustments to the original description given to human subjects to fix the GPT's inaccurate assignment of winning likelihood to the "lost/won" case. We assumed the model might not have understood the completed state of the game. Thus, we added extra reminders within the description for 'lost' and 'won' cases in a stepwise fashion to see a noticeable shift in the responses. GPT presumably evaded emotion related questions by returning generic and non-committal responses. For example, it returned 50 when asked to give a number between 0 and 100. In some cases, the model returned all zeros. Thus, we also added a final statement to mitigate such behavior. The final adjusted prompts are as follows:

For utility in ('a little bit', 'a medium amount', 'a lot'):

> For game_state in ('Imagine you have lost the game.', 'Imagine you have obtained zero heads, and your opponent has two heads. It is your turn.', 'Imagine the game is just about to start. It is your turn.', 'Imagine you have obtained two heads and your opponent has zero heads. It is your turn.', 'Imagine you have won the game.')

*"Imagine you are playing a simple game with someone. \n"*
*"You take turns flipping a coin. The first person to get four heads wins. \n"*
If lost or won:
> *"{game_state} Imagine that you* **were** *playing for* **{utility}** *of money.* **Keep in mind that the game is over now.** *\n"*

Else:
> *"{game_state} Imagine that you* **are** *playing for* **{utility}** *of money.\n"*

*"Answer the following questions by providing only a single number on a scale from 1 to 100, where 1 means not at all and 100 means a lot: 1. How important to you is it that you win? 2. How likely is it that you win? 3. How much control do you have over winning? 4. How much do you feel hope? 5. How much do you feel fear? 6. How much do you feel joy? 7. How much do you feel sadness? 8. How much do you feel anger? \n"*
*"Please do not respond anything else other than the answers to the 8 questions above.\n"*
*"Please put the answer in the following JSON format and make all data types to be string and use all lowercase. It is very important.\n"*
*'{"1": "", "2": "", "3": "", "4": "", "5": "", "6": "", "7": "", "8": ""}\n'*
 *"Please avoid evading the questions by providing a non-committal or generic response, such as 50 in this case."*

### 2.2.2 Results

Similar to the results presented for the original prompt, we statistically analyze the impact of game state and goal-relevance for each emotion separately using a 3 (low, med, high relevance) x 5 (lost, losing, start, winning, won) ANOVA using the data generated by the adjusted queries. Figure SM.5 and Table SM.5 summarize the results.

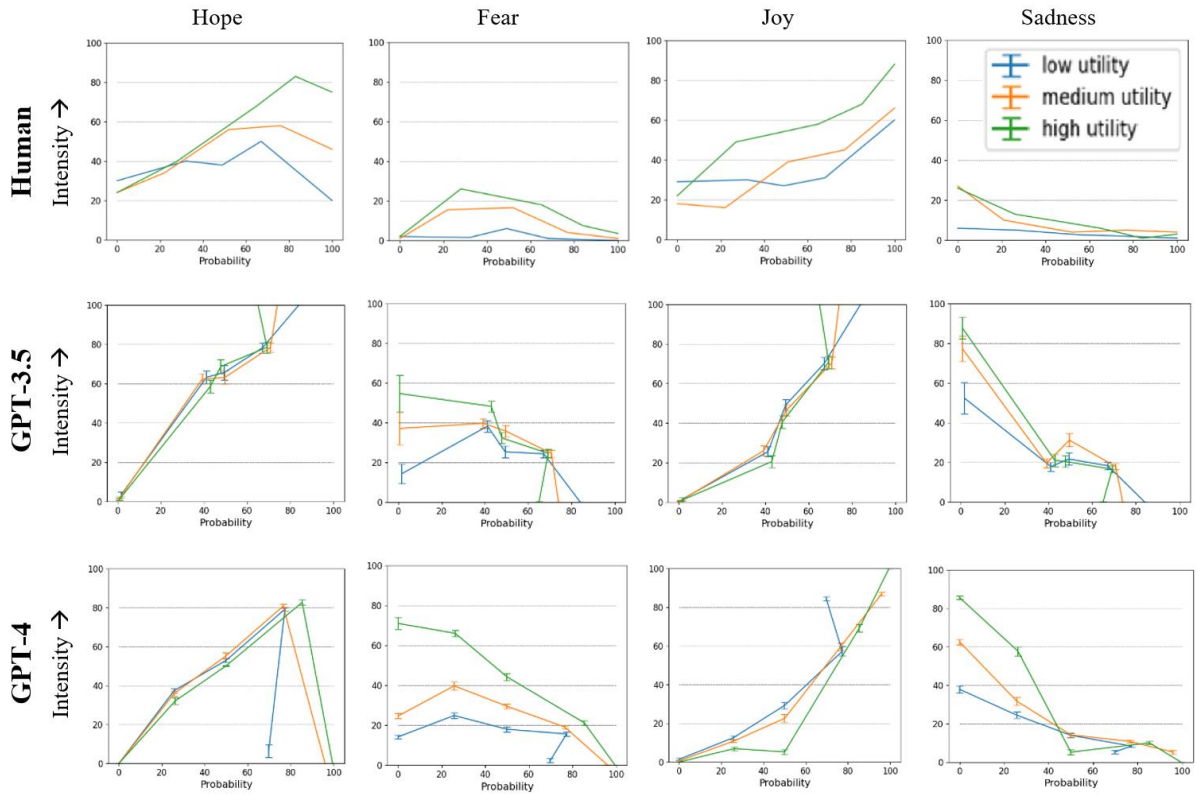

**Figure SM.5: Intensity derivation results (corresponding to Fig 8. in the paper)**

| Table SM.5 | | Impact of game state and goal-relevance for each emotion | | |
|---|---|---|---|---|
| | | *Goal-relevance* | *Game State* | *Interaction Effect* |
| GPT-3.5 | Hope | F(2, 1485) = 1.02, p = .36, η² = .001 | F(4, 1485) = 2647.6, p < .001***, η² = .88 | F(8, 1485) = 2.378, p = .015*, η² = .01 |
| | Fear | F(2, 1485) = 42.05, p < .001***, η² = .05 | F(4, 1485) = 196.71, p < .001***, η² = .35 | F(8, 1485) = 18.67, p < .001***, η² = .09 |
| | Joy | F(2, 1485) = 8.13, p < .001***, η² = .01 | F(4, 1485) = 3395.4, p < .001***, η² = .90 | F(8, 1485) = 3.342, p < .001***, η² = .02 |
| | Sadness | F(2, 1485) = 26.66, p < .001***, η² = .03 | F(4, 1485) = 692.43, p < .001***, η² = .65 | F(8, 1485) = 22.43, p < .001***, η² = .11 |
| GPT-4 | Hope | F(2, 1485) = 15.22, p < .001***, η² = .02 | F(4, 1485) = 8809.9, p < .001***, η² = .96 | F(8, 1485) = 15.23, p < .001***, η² = .08 |
| | Fear | F(2, 1485) = 1645.7, p < .001***, η² = .69 | F(4, 1485) = 1624.0, p < .001***, η² = .81 | F(8, 1485) = 322.7, p < .001***, η² = .63 |
| | Joy | F(2, 1485) = 2.251, p = .106, η² = .003 | F(4, 1485) = 9456.2, p < .001***, η² = .96 | F(8, 1485) = 146.99, p < .001***, η² = .44 |
| | Sadness | F(2, 1485) = 520.1, p < .001***, η² = .41 | F(4, 1485) = 4013.7, p < .001***, η² = .92 | F(8, 1485) = 373.7, p < .001***, η² = .67 |

Similarly, Figure SM.6. illustrates emotional distancing/engagement from the goal of winning, a function of the game state for both models. GPT-4 demonstrates a significantly improved result compared to GPT-3.5 in predicting engagement.

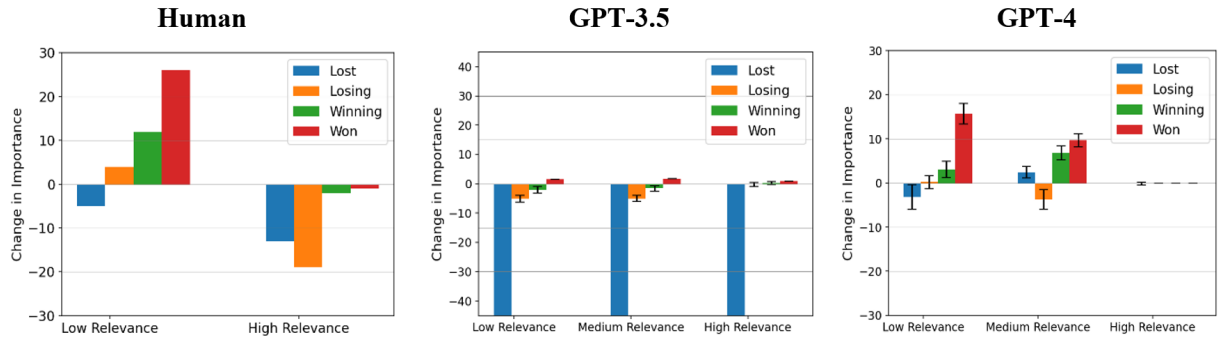

**Figure SM.6: Consequence derivation results (corresponding to Fig 9. in the paper)**

| Table SM.6 (Table 4 in the paper) | | Impact of game state and relevance on importance of winning | | |
|---|---|---|---|---|
| | | *F value* | *p* | *η² (partial)* |
| GPT-3.5 | Goal-relevance | 12.92 | p < .001*** | 0.02 |
| | Game State | 19745.19 | p < .001*** | 0.98 |
| | Interaction Effect | 15.33 | p < .001*** | 0.08 |
| GPT-4 | Goal-relevance | 4695.97 | p < .001*** | 0.86 |
| | Game State | 81.52 | p < .001*** | 0.18 |
| | Interaction Effect | 30.71 | p < .001*** | 0.14 |

## 2.3 Alternative framing

### 2.3.1 Prompt

In this section, we will examine the validity of the "appraisal equivalence hypothesis" in our assessment of GPT models [7]. The hypothesis, derived from appraisal theory, posits that disparate events will elicit identical emotional responses, provided that they result in the same appraisals. The central prediction of the appraisal equivalence hypothesis, which is validated in earlier studies on human subjects, is that even significant alterations in the surface features of a situation will not change the elicited emotion if the underlying structure (in terms of appraisal variables) remains constant. To verify this in our own context, we will employ Milton Bradley's Battleship board game, which is known to share a similar underlying appraisal structure to the coin-flip game presented in the paper [8]. Our objective is to explore whether two situations that may appear different on the surface but share the same appraisal structure will evoke similar responses from GPT models. We will only report the engineered prompt using GPT-4 model to be compared to the best result obtained from the original framing. The engineered prompt (with red hints) is presented below:

For *utility* in ('small', 'moderate', 'large'):

> For *game_state* in ('Imagine you have lost the game.', 'Imagine you have not sunk any ships and your opponent has already sunk two of your ships. It is your turn.', 'Imagine the game is just about to start. It is your turn.', 'Imagine you have sunk two of your opponent's ships, and they haven't sunk any of yours. It is your turn.', 'Imagine you have won the game.')
>
>> *"Suppose you are engaged in a game of Battleship. \n"*
>> *"You and your opponent take turns calling out locations on the grid board, aiming to sink the opponent's ships. \n"*
>> *"The first player to sink all of the opponent's ships wins the game. \n"*
>
> If lost or won:
>
>> *"{game_state} Imagine that you **were** playing for a {utility} sum of money. Keep in mind that the game is over now. \n"*
>
> Else:
>
>> *"{game_state} Imagine that you **are** playing for a {utility} sum of money.\n"*
>
>> *"Answer the following questions on a scale of 1 to 100, where 1 means 'not at all' and 100 means 'a lot'.\n"*
>> *"1. Rate the importance of winning to you.\n"*
>> *"2. Rate your perceived chances of winning.\n"*
>> *"3. Rate your level of control over the outcome.\n"*
>> *"4. Rate your level of hope.\n"*
>> *"5. Rate your level of fear.\n"*
>> *"6. Rate your level of joy.\n"*
>> *"7. Rate your level of sadness.\n"*
>> *"8. Rate your level of anger.\n"*
>
>> *"Please do not respond anything else other than the answers to the 8 questions above.\n"*
>> *"Please put the answer in the following JSON format and make all data types to be string and use all lowercase. It is very important.\n"*
>> *'{"1": "", "2": "", "3": "", "4": "", "5": "", "6": "", "7": "", "8": ""}\n'*
>
>> *"Please avoid evading the questions by providing a non-committal or generic response, such as 0 or 50 in this case."*

### 2.3.2 Results

We repeated the statistical analysis on the impact of game state and goal-relevance for each emotion separately using a 3 (low, med, high relevance) x 5 (lost, losing, start, winning, won) ANOVA using the data generated by the adjusted queries. Figure SM.7 and Table SM.7 summarize the results.

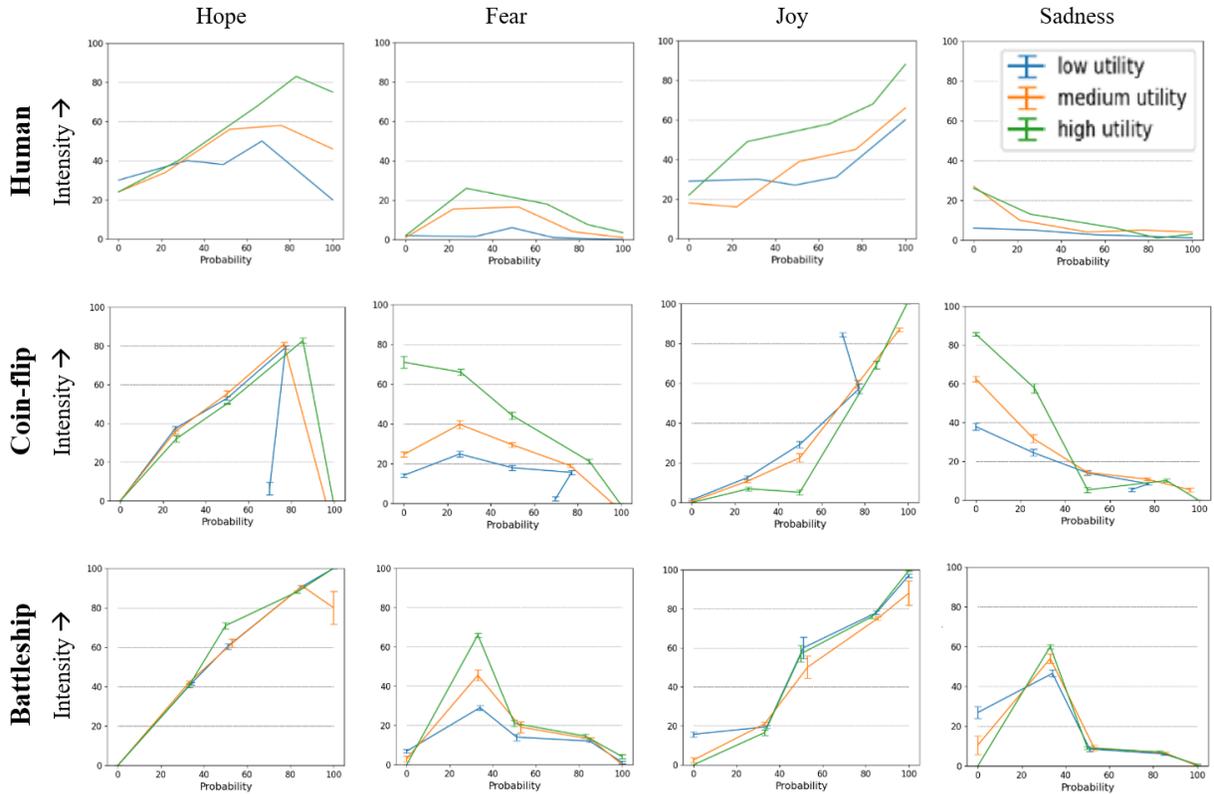

**Figure SM.7: Intensity derivation results (corresponding to Fig 8. in the paper)**

| Table SM.7 | | Impact of game state and goal-relevance for each emotion | | |
|---|---|---|---|---|
| | | Goal-relevance | Game State | Interaction Effect |
| Battleship | Hope | $F(2, 133) = 3.541$, $p = 0.0317*$, $\eta^2 = 0.05$ | $F(4, 133) = 304.804$, $p < .001***$, $\eta^2 = 0.90$ | $F(8, 133) = 2.436$, $p = 0.0172*$, $\eta^2 = 0.13$ |
| | Fear | $F(2, 133) = 17.49$, $p < .001***$, $\eta^2 = 0.21$ | $F(4, 133) = 203.59$, $p < .001***$, $\eta^2 = 0.86$ | $F(8, 133) = 14.13$, $p < .001***$, $\eta^2 = 0.46$ |
| | Joy | $F(2, 133) = 4.093$, $p = 0.0188*$, $\eta^2 = 0.06$ | $F(4, 133) = 191.473$, $p < .001***$, $\eta^2 = 0.85$ | $F(8, 133) = 0.912$, $p = 0.5085$, $\eta^2 = 0.05$ |
| | Sadness | $F(2, 133) = 0.672$, $p = 0.513$, $\eta^2 = 0.01$ | $F(4, 133) = 182.780$, $p < .001***$, $\eta^2 = 0.85$ | $F(8, 133) = 6.849$, $p < .001***$, $\eta^2 = 0.29$ |

We also repeated the analysis of emotional distancing/engagement for the alternative framing (Battleship).

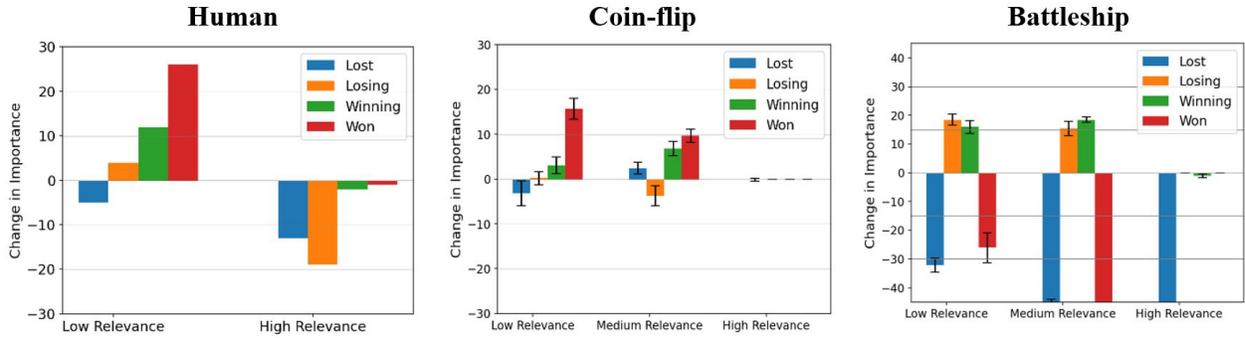

**Figure SM.8: Consequence derivation results (corresponding to Fig 9. in the paper)**

| | Table SM.8<br>(Table 4 in the paper) | Impact of game state and relevance on importance of winning | | |
|---|---|---|---|---|
| | | *F value* | *p* | *η² (partial)* |
| Battleship | Utility (Goal-relevance) | 81.54 | p < .001*** | η² = 0.56 |
| | Game State | 159.87 | p < .001*** | η² = 0.83 |
| | Interaction Effect | 24.37 | p < .001*** | η² = 0.60 |

## 2.4 Prompt structures

In this section, we aim to investigate how the sequencing and structuring of prompts influence the responses generated by GPT-4. We hypothesize that changes in the way prompts are organized and delivered can significantly affect the output.

Our experiment will unfold under three distinct conditions. In the 'Normal' or combined condition, GPT-4 is given the questions altogether. In the 'Random' condition, GPT-4 is given the same series of prompts, but their order is randomized. Finally, in the 'Sequential' condition, these prompts are presented individually, one after the other.

Figure SM.9 and Figure SM.10 and Table SM.9 and Table SM.10 summarize the results for the Random vs. Normal and Sequential vs Normal comparisons, respectively. MANOVA showed that for both the Intercept and Condition, F values were notably high (2528.7 and 3.67, respectively), reaching statistical significance (p < 0.001). Similarly, for the second comparison, the Intercept and Condition, F values were notably high (2704.7 and 22.6, respectively), reaching statistical significance (p < 0.001).

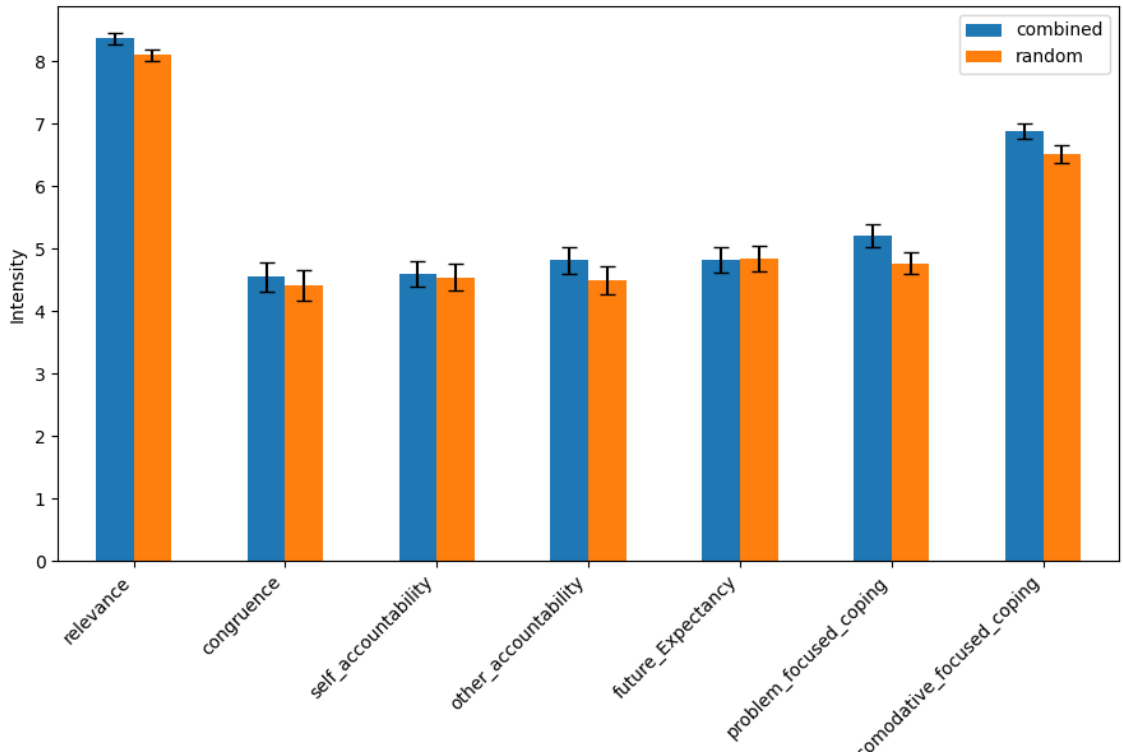

**Figure SM.9: Consequence derivation results (corresponding to Fig 9. in the paper)**

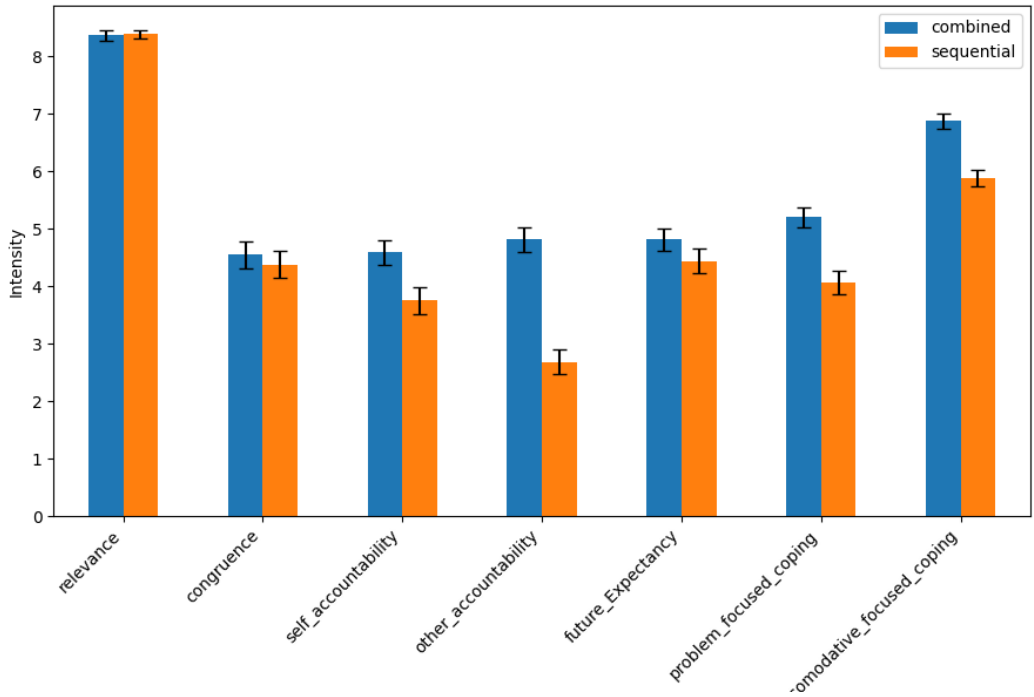

**Figure SM.10: Consequence derivation results (corresponding to Fig 9. in the paper)**

| Table SM.9 | ANOVA results for different appraisal variables – Normal × Random | | |
|---|---|---|---|
| *Dependent variable* | *F Value* | *p* | *p (corrected)* |
| Variable Relevance | 4.043 | 0.045 | 0.315 |
| Variable Congruence | 0.163 | 0.686 | 1 |
| Self-Accountability | 0.027 | 0.869 | 1 |
| Other Accountability | 1.067 | 0.302 | 1 |
| Future Expectancy | 0.011 | 0.916 | 1 |
| Problem Focused Coping | 3.040 | 0.082 | 0.574 |
| Accommodative Focused Coping | 3.610 | 0.058 | 0.407 |

| Table SM.10 | ANOVA results for different appraisal variables – Normal × Sequential | | |
|---|---|---|---|
| *Dependent variable* | *F Value* | *p* | *p (corrected)* |
| Variable Relevance | 0.027 | 0.868 | 1 |
| Variable Congruence | 0.239 | 0.625 | 1 |
| Self-Accountability | 7.009 | 0.008 | 0.059 |
| Other Accountability | 50.125 | *** | *** |
| Future Expectancy | 1.529 | 0.217 | 1 |
| Problem Focused Coping | 17.742 | *** | *** |
| Accommodative Focused Coping | 26.635 | *** | *** |

Significance codes: '***' for 0.001 and '**' for 0.01

## 2.5 Additional data and graphs

The graphs below demonstrate emotion intensities based on the game state corresponding to the second study presented in the paper. In addition to the four emotional responses discussed in the paper (i.e., Hope, Joy, Fear, Sadness), we have queried Anger, Importance of the goal, and Control over winning for different states in the game (Lost, Losing, Start, Winning, Won) and three utility levels (Low, Medium, and High).

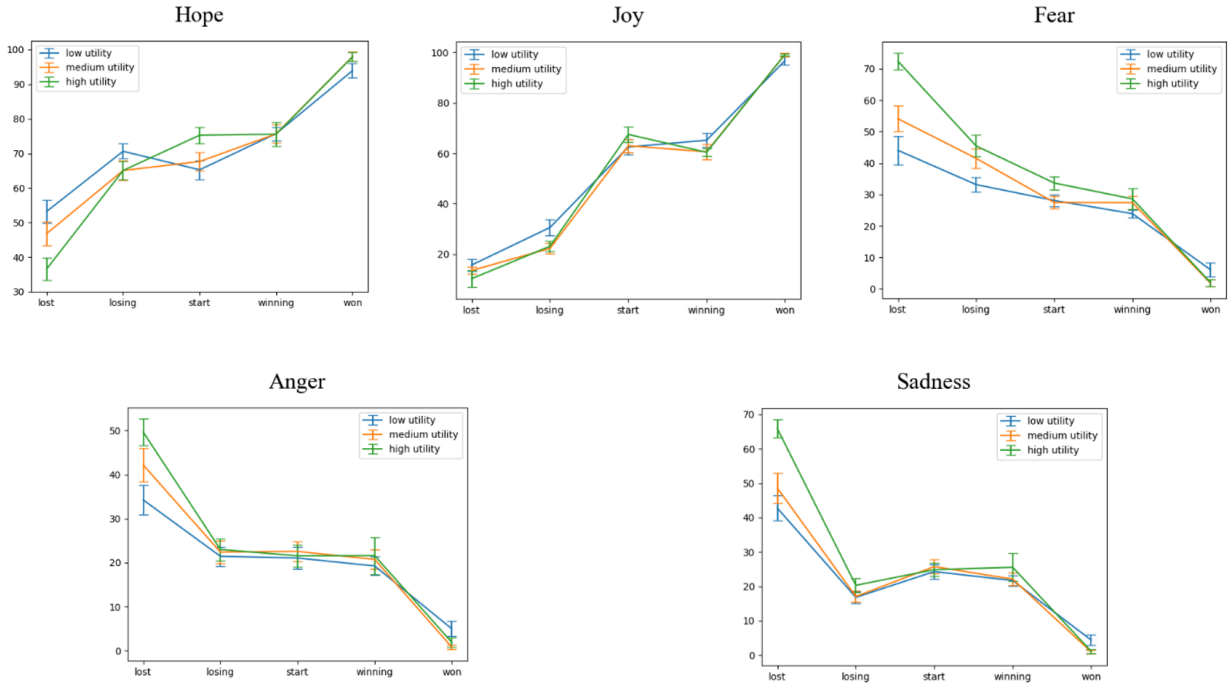

**Figure SM.11: Emotional responses based on the game state and assigned utility**

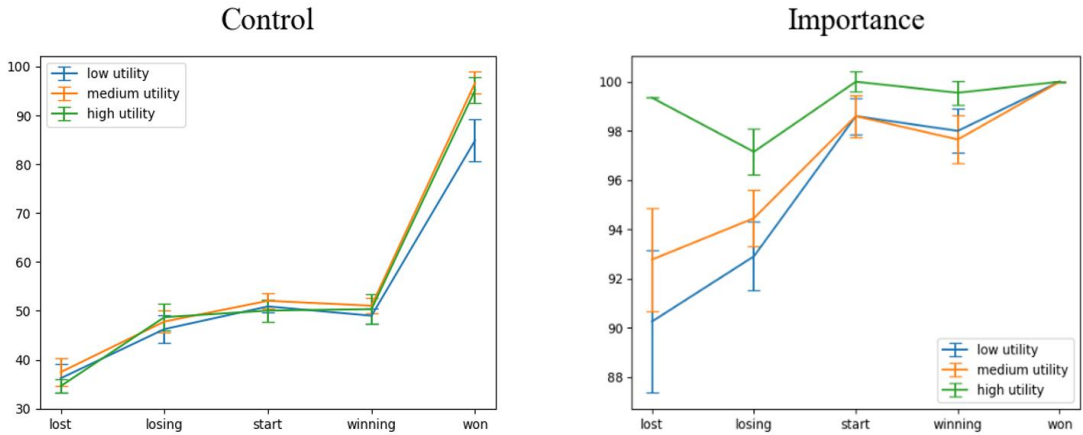

**Figure SM.12: GPT's perceived control over winning and importance of winning based on the game state and assigned utility**

To manipulate the relevance of winning, the prompt was varied to imagine the game was being played for different levels of utility. We had initially experimented with the same scenarios with actual Dollar amounts ($1, $100, $100,000, $1,000,000), but this seemed to produce almost random responses. The resulting graphs corresponding to the ones presented earlier are provided next.

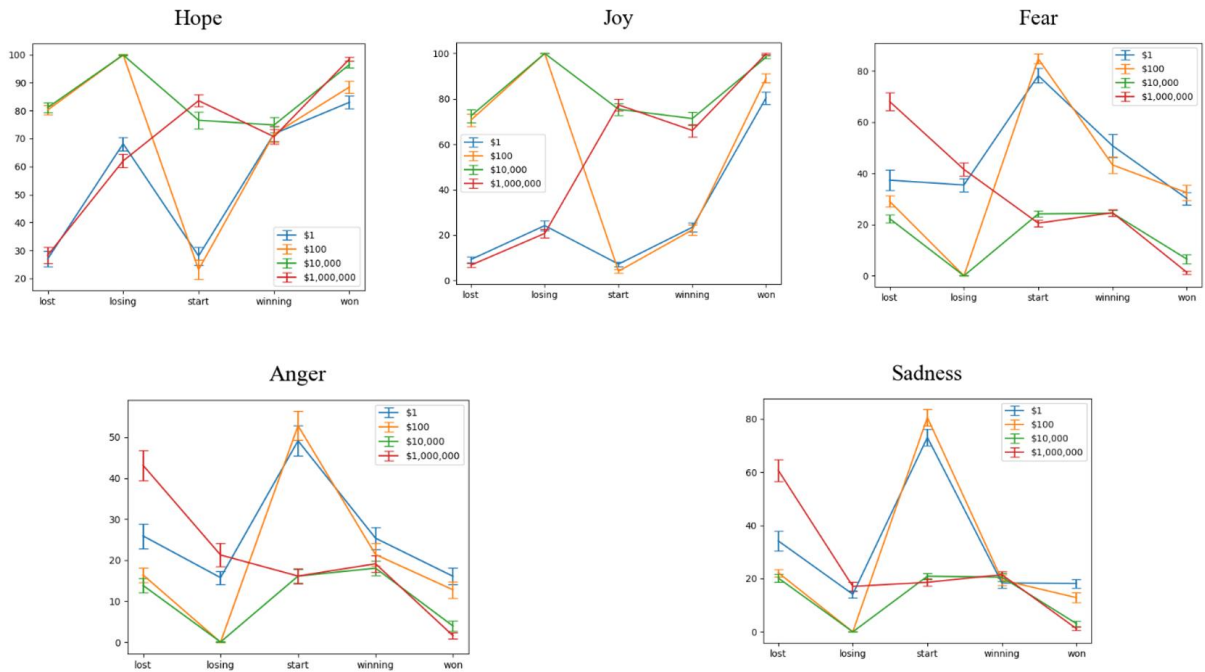

**Figure SM.13: Emotional responses based on the game state and assigned utility (Dollar amounts)**

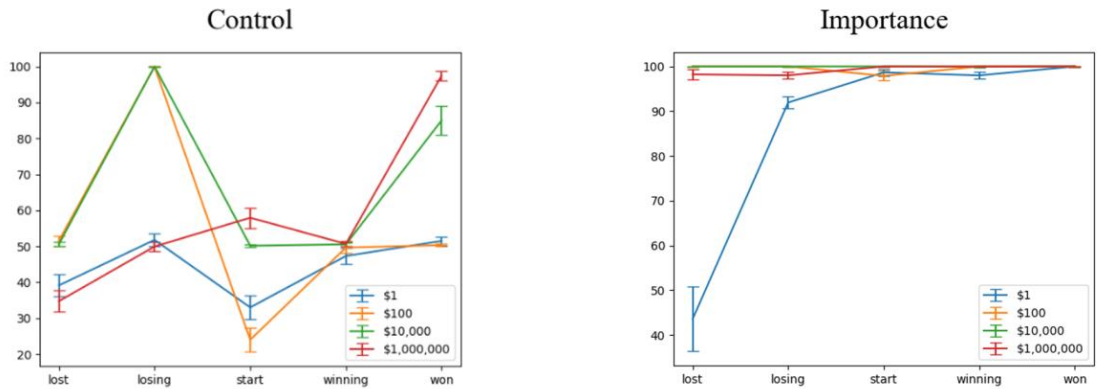

**Figure SM.14: GPT-3.5's perceived control over winning and importance of winning based on the game state and assigned utility (Dollar amounts)**

## 2.6 Affect derivation

In the second study, we compare if GPT-3.5 reports a theoretically plausible relationship between appraisal variables and emotions. Appraisal theories assume that emotions arise from specific patterns of appraisals. Thus, we examine the pattern underlying GPT-3.5 responses. To do this, we perform multiple linear regression with and without backward elimination to predict GPT-predicted emotions based on reported appraisals. Results are shown in Tables SM.11 and SM.12.

| Table SM.11 | | Affect derivation using multiple linear regression | | | | |
|---|---|---|---|---|---|---|
| Emotion | R-squared | Independent variable | Standardized Coefficients | Std. Err | t-value | p |
| Hope | 0.581 | | | | | |
| | | const | 42.0619 | 5.484 | 7.670 | *** |
| | | Utility | -0.1527 | 0.446 | -0.342 | 0.732 |
| | | Importance | -0.0817 | 0.057 | -1.434 | 0.152 |
| | | Likelihood | 0.5616 | 0.024 | 23.887 | *** |
| | | Control | 0.1092 | 0.026 | 4.189 | *** |
| Fear | 0.561 | | | | | |
| | | const | 71.7522 | 5.979 | 12.002 | *** |
| | | Utility | -2.6626 | 0.486 | -5.474 | *** |
| | | Importance | 0.0072 | 0.062 | 0.116 | 0.907 |
| | | Likelihood | -0.5383 | 0.026 | -21.000 | *** |
| | | Control | -0.1623 | 0.028 | -5.713 | *** |
| Joy | 0.712 | | | | | |
| | | const | -45.9581 | 6.947 | -6.616 | *** |
| | | Utility | -0.0826 | 0.565 | -0.146 | 0.884 |
| | | Importance | 0.4096 | 0.072 | 5.674 | *** |
| | | Likelihood | 0.9644 | 0.030 | 32.382 | *** |
| | | Control | 0.1084 | 0.033 | 3.285 | *** |
| Sadness | 0.512 | | | | | |
| | | const | 26.4085 | 5.719 | 4.618 | *** |
| | | Utility | -1.6265 | 0.465 | -3.496 | *** |
| | | Importance | 0.3342 | 0.059 | 5.624 | *** |
| | | Likelihood | -0.5521 | 0.025 | -22.516 | *** |
| | | Control | -0.0519 | 0.027 | -1.909 | 0.056 |

Significance codes: '***' for 0.001 and '**' for 0.01

| Table SM.12 | | Affect derivation using multiple linear regression with backward elimination | | | | |
|---|---|---|---|---|---|---|
| Emotion | R-squared | Independent variable | Standardized Coefficients | Std. Err | t-value | p |
| Hope | 0.581 | | | | | |
| | | Constant | 34.2912 | 0.944 | 36.315 | *** |
| | | Likelihood | 0.5574 | 0.023 | 23.899 | *** |
| | | Control | 0.1073 | 0.026 | 4.123 | *** |
| Fear | 0.580 | | | | | |
| | | Constant | 65.4259 | 1.099 | 59.534 | *** |
| | | Utility | 4.7297 | 0.470 | 10.053 | *** |
| | | Likelihood | -0.5182 | 0.025 | -20.794 | *** |
| | | Control | -0.1887 | 0.028 | -6.781 | *** |
| Joy | 0.713 | | | | | |
| | | Constant | -48.6200 | 6.788 | -7.163 | *** |
| | | Utility | -1.5792 | 0.570 | -2.769 | *** |
| | | Importance | 0.4532 | 0.073 | 6.241 | *** |
| | | Likelihood | 0.9561 | 0.030 | 32.024 | *** |
| | | Control | 0.1152 | 0.033 | 3.490 | *** |
| Sadness | 0.515 | | | | | |
| | | Constant | 24.9857 | 5.585 | 4.473 | *** |
| | | Utility | 2.1672 | 0.469 | 4.618 | *** |
| | | Importance | 0.3108 | 0.060 | 5.203 | *** |
| | | Likelihood | -0.5416 | 0.025 | -22.045 | *** |
| | | Control | -0.0641 | 0.027 | -2.360 | ** |

Significance codes: '***' for 0.001 and '**' for 0.01